\documentclass[conference]{IEEEtran}
\usepackage{times}

\usepackage[numbers]{natbib}
\usepackage{multicol}
\usepackage[bookmarks=true]{hyperref}
\usepackage{graphicx}
\usepackage{tabularx}
\usepackage{amssymb}
\usepackage{amsmath}
\usepackage{fancyhdr}
\usepackage{algorithmic}
\usepackage{tikz}
\usepackage{bm}
\usepackage{multicol}
\usepackage{listings}

\algsetup{linenosize=\small}
\usepackage{algorithm}
\newcommand{\matr}[1]{\mathbf{#1}} 

\newcommand{\vect}[1]{\mathbf{#1}} 
\newcommand{\vectg}[1]{\boldsymbol{#1}}

\newcommand{\bx}{\mathbf{x}}
\newcommand{\bu}{\mathbf{u}}
\newcommand{\bff}{\mathbf{f}}
\newcommand{\bR}{\mathbf{R}}

\newcommand{\bex}{{\bf e}_{x}}

\newcommand{\bezr}{{\bf e}_{z_r}}
\newcommand{\bexd}{{\bf e}_{x_s}}
\newcommand{\beyd}{{\bf e}_{y_s}}

\newcommand{\norm}[1]{{\|{#1}\|}}

\begin{document}
\begin{titlepage}
\vspace*{\fill}
{\large
\copyright 2020 IEEE.  Personal use of this material is permitted.  Permission from IEEE must be obtained for all other uses, in any current or future media, including reprinting/republishing this material for advertising or promotional purposes, creating new collective works, for resale or redistribution to servers or lists, or reuse of any copyrighted component of this work in other works}
\vspace*{\fill}
\end{titlepage}
\title{\vspace{-0.75cm} Direct NMPC for Post-Stall Motion Planning with Fixed-Wing UAVs}

\author{Max Basescu and Joseph Moore}


%

\maketitle

\begin{abstract}
Fixed-wing unmanned aerial vehicles (UAVs) offer significant performance advantages over rotary-wing UAVs in terms of speed, endurance, and efficiency. However, these vehicles have traditionally been severely limited with regards to maneuverability. In this paper, we present a nonlinear control approach for enabling aerobatic fixed-wing UAVs to maneuver in constrained spaces. Our approach utilizes full-state direct trajectory optimization and a minimalistic, but representative, nonlinear aircraft model to plan aggressive fixed-wing trajectories in real-time at 5 Hz across high angles-of-attack. Randomized motion planning is used to avoid local minima and local-linear feedback is used to compensate for model inaccuracies between updates. We demonstrate our method in hardware and show that both local-linear feedback and re-planning are necessary for successful navigation of a complex environment in the presence of model uncertainty. 
\end{abstract}

\IEEEpeerreviewmaketitle

\section{Introduction}

Fixed-wing unmanned aerial vehicles (UAVs) have long been viewed as valuable assets in both defense and commercial domains. Their endurance and speed makes them ideal for gathering information or transporting cargo across open sky. However, for robotics applications where navigation around obstacles is a necessity, fixed-wings have typically been overlooked in favor of rotary-wing vehicles. Quadcopter UAVs in particular, with their simple mechanical design and high thrust-to-weight ratios, have often been preferred for these applications. This is predominantly because at low speeds and in linear regimes, quadcopters can achieve near-zero turn radii. Fixed-wings, in contrast, have a fixed (often sizable) minimum turn radius in traditional steady-level flight regimes.

At high speeds and in more dynamic regimes, simplified differentially-flat quadcopter models exist which dramatically reduce the computation time needed to generate feasible trajectories \cite{mellinger2011minimum}. These simplified models have been shown to work well in practice \cite{mellinger2012trajectory}. Such models do not exist for fixed-wings traveling at high angles-of-attack, however high angles-of-attack regimes are necessary for fixed-wings to achieve tight turn radii (see Figure \ref{Fig:turnradii}).

In this paper, we present a nonlinear model predictive control approach which uses direct trajectory optimization (i.e., direct NMPC) to compute dynamically feasible, collision-free trajectories in real-time. Our approach hinges on a minimalistic dynamics model and solving a very fast feasibility problem, rather than an optimization problem, while enforcing state and actuator constraints. We also employ local linear feedback to compensate for modeling errors between re-plans. We first demonstrate that post-stall aerobatics provide a significant advantage in terms of minimum achievable turn-radii for fixed-wing UAVs. We then present our simplified 17-state model to be used by the optimizer and use this model to conduct a performance evaluation of the trajectory optimizer as a function of integration method and number of knot points. Finally, we conduct a series of hardware experiments where we evaluate our approach. We compare the performance of our algorithm with and without local feedback between plans.

\label{sec:intro}
\section{Post-Stall Motion Planning}

\begin{figure}
  \centering
  \includegraphics[width=.95\linewidth]{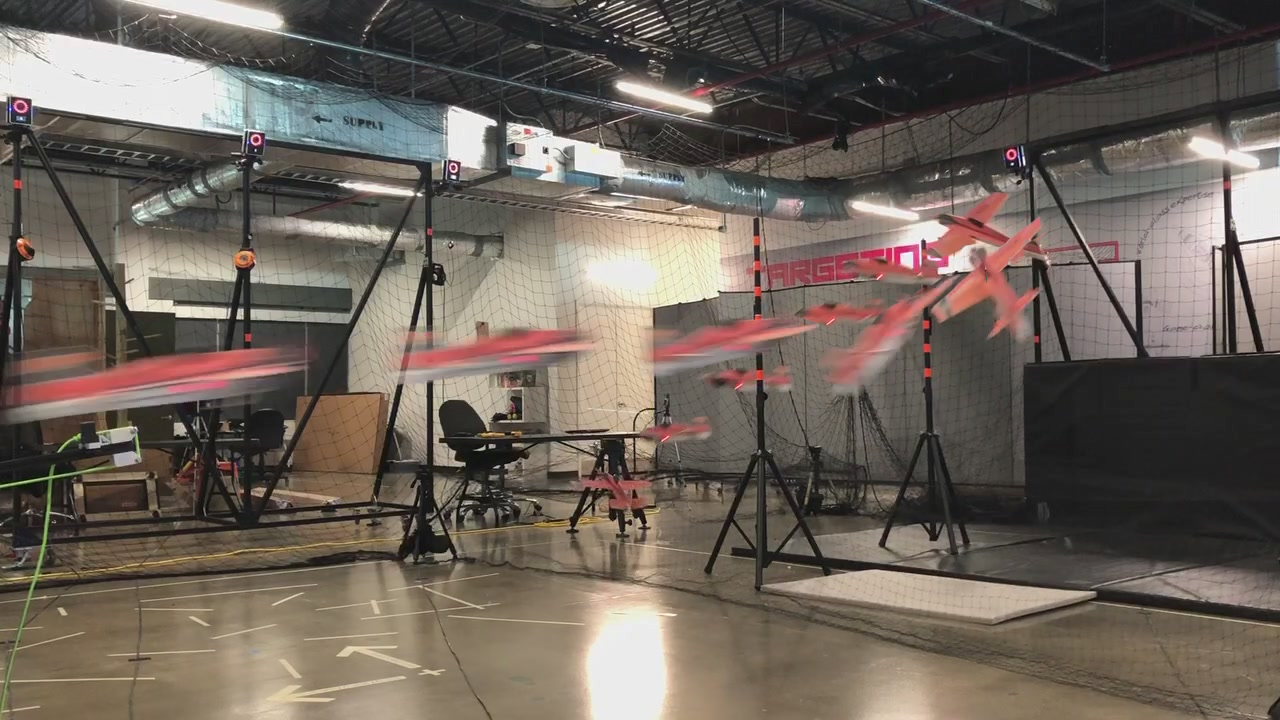}
\caption[Aerobatic aircraft executing a post-stall 90 degree turn.]{Aerobatic aircraft executing a post-stall 90 degree turn through a virtual hallway denoted by two striped poles.}
  \label{Fig:timelapse}
  \end{figure}

Early work in aircraft design and control recognized the value of post-stall maneuvers for increasing aircraft maneuverability \cite{herbst1984supermaneuverability}. However, in combat, these high angle-of-attack maneuvers are often viewed as impractical, since they make an aircraft vulnerable by reducing its kinetic energy \cite{kutschera2000performance}. 

\begin{figure}
  \centering
  \includegraphics[width=.9\linewidth]{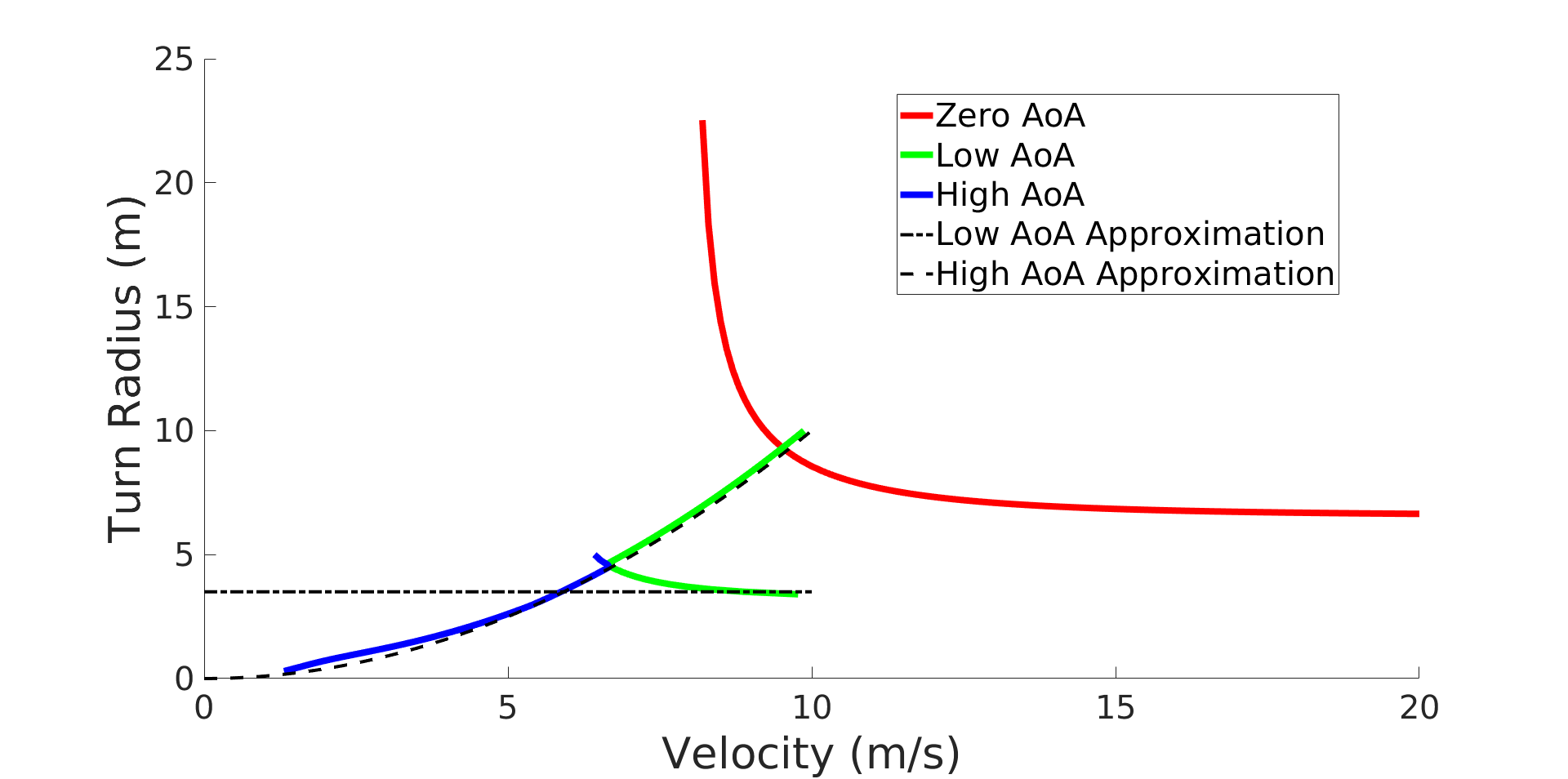}
\caption[Turn radius.]{Plot demonstrates how fixed-wing UAVs achieve smaller turn radii by executing high angle-of-attack maneuvers. Curves were generated by computing turn trim conditions using a nonlinear program and a simplified nonlinear aircraft model without control surfaces. Constraints were placed on angle-of-attack to demonstrate the advantage of post-stall turns.}
  \label{Fig:turnradii}
  \end{figure}

In the context of mobile robotics, however, these post-stall maneuvers have the potential to dramatically reduce radius of curvature, allowing a fixed-wing vehicle to navigate through very constrained environments (see Figure \ref{Fig:turnradii}). We believe that the tight turning radii capable of being achieved by fixed-wing UAVs at high angles-of-attack put them on par with quadcopter maneuverability and motivates the need for post-stall kinodynamic motion planning strategies. 
\label{sec:intro}
\section{Related Work}

In comparison to rotorcraft UAVs \cite{liew2017recent}, significantly less work has been done to explore the use of fixed-wing UAVs as a platform for mobile robotics. Some of the earliest and most notable work in post-stall maneuvers with fixed-wing UAVs is presented in \cite{sobolic2009nonlinear}, where researchers explored transition to-and-from a prop-hang configuration. This was preceded by work in \cite{milam2002real}, which sought to generate real-time trajectories for a planar ducted-fan UAV with aerodynamic surfaces. The control of fixed-wing UAVs in post-stall flight was also advanced through the exploration of post-stall perching \cite{moore2014robust}, where a library of trajectories was used to enable robust perching performance. Motion planning for fixed-wing UAVs was also explored in \cite{majumdar2016funnel}, where a library of funnels exploited invariance of initial conditions to navigate safely through an obstacle field in real-time. The initial work was limited to simple planar model representations, but then was extended to higher dimensional models which could exploit the full angle-of-attack envelope. More recently, researchers have explored real-time trajectory motion-planning of fixed-wing UAVs using simplified models restricted to low angle-of-attack domains \cite{alturbeh2014real}. Authors have also investigated in simulation aggressive aerobatic fixed-wing turn-around maneuvers \cite{matsumoto2010agile,levin2016aggressive}. The authors of \cite{levin2017agile, khan2016modeling, bulka2018autonomous} have for the last several years been leading the way in developing control and planning strategies for aerobatic fixed-wing vehicles. Their work has been characterized by very high fidelity physics models \cite{khan2016modeling} which break aerodynamic surfaces into segments, capture transient aerodynamic effects, and model motor and propeller physics. Several recent publications \cite{levin2019real, bulka2019high} have designed real-time motion planners using trajectory libraries generated offline using these high fidelity physics models.

The work presented here takes a different approach. Instead of building a high-fidelity physics model of the aircraft, we build a medium-fidelity physics model which is amenable to real-time trajectory optimization. Despite the computational complexity \cite{dekka2016improved}, we choose to utilize direct trajectory optimization methods because of their improved numerical conditioning, sparse constraints, and ability of the gradient computations to be trivially parallelized \cite{tedrake2009underactuated}. As discussed in \cite{levin2017agile}, these direct methods can also be easily seeded by a simplified randomized motion planner. We formulate a nonlinear feasibility problem which can be solved at real-time rates and use local linear feedback to stabilize the vehicle between replans. We also optimize and control over the entire state-space, so as to not make assumptions of timescale separations. 

\label{sec:intro}
\section{Approach}
\label{sec:approach}
We believe that to navigate in complex environments which are unknown a-priori, we must be able to generate sophisticated trajectories in real-time that exploit the full flight envelope for fixed-wing vehicles and achieve ``supermaneuverability'' \cite{herbst1984supermaneuverability}. Trajectory libraries, while commonly used to reduce online computational burdens, often become computationally expensive and result in sub-optimal solutions when the vehicle state space is large ($>$6) and the environment is complex \cite{milam2000new}. In addition, our approach must be able to handle complicated physics models with complex aerodynamics, which are in general not differentially flat.

Our approach utilizes a minimalistic nonlinear model and real-time direct trajectory optimization to rapidly solve a feasibility problem in real-time. In comparison to other real-time trajectory generation approaches, we believe that our direct method has several advantages, including the ability to exploit problem sparsity and impose state constraints. 
\subsection{Dynamics Model}

We extend the planar post-stall aircraft model presented in \cite{moore2014robust} into a three-dimensional model of a fixed-wing UAV with a propeller in a ``puller'' configuration. 
We define our state as $\vect{x}=\begin{bmatrix} x_r,y_r,z_r,\phi,\theta,\psi,\delta_{ar},\delta_{al},\delta_e,\delta_r,\delta_t,v_{x},v_{y},v_{z},\omega_x,\omega_y,\omega_z\end{bmatrix}^T$.
Here $\vect{r} =\begin{bmatrix} x_r,y_r,z_r\end{bmatrix}^T$ represents the position of the center of mass in the world frame $O_{x_r y_r z_r}$, $\vectg{\theta}= \begin{bmatrix} \phi, \theta, \psi\end{bmatrix}^T$ represents the set of $z$-$y$-$x$ euler angles, $\vectg{\delta} = \begin{bmatrix} \delta_{ar}, \delta_{al}, \delta_e, \delta_r \end{bmatrix}^T$ are control surface deflections due to the right and left ailerons, elevator and rudder respectively, $\delta_t$ is the magnitude of thrust from the propeller, $\vect{v} = \begin{bmatrix} v_x, v_y, v_z\end{bmatrix}^T$ is the velocity of the center of mass in the \emph{body}-fixed frame $O_{xyz}$, and $\vectg{\omega} = \begin{bmatrix}\omega_x, \omega_y, \omega_z\end{bmatrix}^T$ represents the angular velocity of the body in the body-fixed frame. We can then write $\vect{x} = \begin{bmatrix} \vect{r}, \vectg{\theta}, \vectg{\delta}, \delta_t, \vect{v}, \vectg{\omega}\end{bmatrix}^T$,  $\vect{u_{cs}} = \begin{bmatrix} \omega_{ar}, \omega_{al}, \omega_{e}, \omega_{r}, u_t\end{bmatrix}^T$.
The equations of motion then become
\begin{align}
\dot{\vect{r}} &= \bR_b^r \vect{v},\quad \dot{\vectg{\theta}} = {\bR}_{\omega}^{-1}{\boldsymbol \omega}\nonumber\\
\dot{\vectg{\delta}} &= \vect{u}_{cs}, \quad \dot{\delta}_t = a_t \delta_t + b_t u_t\nonumber\\
\dot{\vect{v}} &= \vect{f}/m - \vectg{\omega}\times \vect{v},\quad \dot{\vectg{\omega}} = \matr{J}^{-1}(\vect{m}-\vectg{\omega}\times\matr{J}\vectg{\omega})
\end{align}
Here $m$ is vehicle mass, $\matr{J}$ is the vehicle's inertia tensor with respect to the center of mass, $\vect{f}$ are the total forces applied to the vehicle in body-fixed coordinates, $\vect{m}$ are the moments applied about the vehicle's center of mass in body-fixed coordinates, $\vect{u}_{cs}$ are the angular velocity inputs of the control surfaces.
$\bR_b^r$ denotes the rotation from the body-fixed frame to the world frame, and ${\bR}_{\vectg{\omega}}$ is the rotation which maps the euler angle rates to an angular velocity in body-fixed frame.

The forces acting on the vehicle can be written as
\begin{align}
\vect{f}= \sum_i \bR_{s_i}^b \vect{f}_{s_i} + {\bR_b^r}^T g\bezr + \sum_i {\bR_{t}^b}\vect{f}_{t}
\end{align}
where $\vect{f}_{s_i}$ represent the forces due to the aerodynamic surfaces, $\vect{f}_{t}$ represents the forces due to the propeller and is given as $\begin{bmatrix}\delta_t & 0 & 0\end{bmatrix}^T$. $\bR_{t_i}^b$ is the rotation matrix that defines the orientation of the thrust source with respect to the body fixed frame. $\bR_{s_i}^b$ is the rotation matrix that defines the aerodynamic surface reference frame with respect to the body fixed frame. The aerodynamics surfaces used for this model are the wing, the horizontal and vertical fuselage components, the horizontal and vertical tail, and the control surfaces. 

To model forces on the aerodynamic surfaces, we use
\begin{align}
\vect{f}_{s}=f_{n,s_i}\beyd = \frac{1}{2}C_{n}\rho|\vect{v}_{s_i}|^2 S_i\beyd
\end{align}
where
$\vect{v}_{s_i}$ is the velocity of the $i^{th}$ aerodynamic surface in the surface reference frame given as
\begin{align}
\vect{v}_{s _i}&=\bR_{s_i}(\vect{v}+\vectg{\omega}\times\vect{r}_h + \gamma_i \vect{v}_{bw}) + (\bR_{s_i}\vectg{\omega}+\vectg{\omega}_{s_i})\times\vect{r}_{s_i}.
\end{align}
Here $\vect{r}_h$ represents the displacement from the vehicle center of mass to the aerodynamic surface ``hinge'' point, $\bR_{s_i}$ represents the rotation matrix that transforms vectors in the body-fixed frame into the aerodynamic surface frame, $\vectg{\omega}_{s_i}$ is the simple rotation rate of the aerodynamic surface in the aerodynamic surface frame. This is only non-zero for actuated surfaces. 

$\vect{v}_{bw}$ is the velocity due to the backwash of the propeller. It can be approximated using momentum theory as,
\begin{align}
\vect{v}_{bw}= v_{bw}\bex = \sqrt{\norm{\vect{v}_p}_2^2+\frac{2\delta_{t}}{\rho S_{disk}}} -\norm{\vect{v}_p}_2.
\end{align}
where $\delta_t$ is the thrust input, $S_{disk}$ is the area of the actuator disk, $\rho$ is the density of air and $\vect{v}_p$ is the velocity at the propeller. $\gamma$ is an empirically determined backwash velocity coefficient.
For a given aerodynamic surface, the angle of attack becomes
\begin{align}
\alpha_{s}= \arctan{\frac{v_{s,z}}{v_{s,x}}}
\end{align}

$\vect{m}$ in the body fixed frame can be given as
\begin{align}
\vect{m}= \sum_{i}(\vect{r}_{s_i} \times \bR_{s}\vect{f}_{s_i}),
\end{align}
where $\vect{r}_{s_i}=l_h\bex+\bR_{s}(l_{s}\bexd)$.
To model the thrust dynamics, we assume a first order linear model for the thrust given above
where $a_t$ and $b_t$ are constants and $u_t$ is the normalized PWM value, where $u_t\in{[0,1]} $. 
$C_{n}$ comes from the flat plate model in \cite{hoerner1985fluid} and is given as $C_{n}=2\sin\alpha_w$.

\subsection{Control Strategy}

To achieve real-time planning for fixed-wing vehicles across the entire flight envelope, we utilize a four-stage, hierarchical control strategy consisting of a randomized motion planner, a spline-based smoothing algorithm, direct trajectory optimization, and local linear feedback control. The spline-based smoothed path is used as a seed to the trajectory optimizer and also provides a receding horizon goal point. The local linear feedback controller compensates for model-uncertainty and environmental disturbances between plans.


%

\subsubsection{RRT Generation and Spline-Based Smoothing}
The seed path is generated through a combination of standard RRT \cite{Lavalle:1998:Rapidly}, spline-based smoothing satisfying a maximum curvature constraint \cite{yang2010analytical}, and a final re-parametrization of the trajectory.  This combination of steps is useful for obtaining a path which is more dynamically feasible than a raw RRT, and also helps to intelligently select an endpoint for the trajectory optimizer.


First, the RRT is grown in a three dimensional state space with a 10\% bias towards the goal point, until it arrives within an error ball around the goal.  The resultant path $\vect{x} = [x_0,x_1,\ldots,x_n]$ from start to goal is then pruned as described in \cite{yang20083d} to remove extraneous nodes.

We then employ G2 Continuous Cubic B\'ezier Spiral Path Smoothing (G2CBS) to obtain a continuous curvature, smooth path \cite{yang2010analytical}.  This algorithm examines each set of three adjacent nodes $[W_1,W_2,W_3]$ in the pruned path.  It transforms them into a local 2D space, computes control points for two symmetric G2 splines satisfying a maximum curvature constraint $\kappa_{max}$ (set to $2m^{-1}$), then transforms the control points back into 3D.  The output of this algorithm is a set of 3D control points for each spline,
\begin{align}
\matr{B_c} &= \begin{bmatrix}
\vect{B_0}&\vect{B_1}&\vect{B_2}&\vect{B_3}
\end{bmatrix}\in \mathbb{R}^{3x4}\nonumber\\
\matr{E_c} &= \begin{bmatrix}
\vect{E_3}&\vect{E_2}&\vect{E_1}&\vect{E_0}
\end{bmatrix}\in \mathbb{R}^{3x4}
\end{align}
A 3rd order B\'ezier curve can be evaluated by,
\begin{align}
\vect{F}(s)&=(1-s)^{3}\vect{P_0}+3(1-s)^{2}s\vect{P_1}\ldots\nonumber\\&+3(1-s)s^{2}\vect{P_2}+s^{3}\vect{P_3},\ s\in[0,1]
\end{align}
where $[\vect{P_0},\ldots,\vect{P_3}]$ is the set of control points for the spline.
This expression allows us to evaluate the $B$ and $E$ splines represented by the control points $\matr{B_c}$ and $\matr{E_c}$, and to stitch together the overall G2 continuous path $\vect{x}$, which consists of straight line segments and spline segments.  

We extend this approach to compute the curvature of the entire path, in order to apply a rough kinematic mapping between curvature and velocity.  This mapping allows us to reparametrize the seed path in terms of time.  By picking an endpoint for the trajectory optimizer based on a constant time along the seed path, the duration of the optimized trajectory is somewhat regularized.  This helps to prevent the solver from wasting computational resources planning too far into the future, and allows us to provide a more feasible estimate of final velocity to the solver.  We compute the first and second order derivatives of $\vect{F}$ with respect to $s$, $\vect{F'}$ and $\vect{F''}$.
Pulling out the individual elements as,
\begin{align}
\vect{F'}(s)&=\begin{bmatrix}x'&y'&z'\end{bmatrix},\ \vect{F''}(s)=\begin{bmatrix}x''& y''&z''\end{bmatrix}
\end{align}
We can express the curvature along $\vect{F}(s)$ as,
\begin{align}
\kappa(s) = \frac{\sqrt{(z''y'-y''z')^2+(x''z'-z''x')^2+(y''x'-x''y')^2}}{(x'^2+y'^2+z'^2)^{\frac{3}{2}}}
\end{align}
with the curvature along the straight segments of path equal to 0.  The curvature along the entire path is then mapped to velocity as,
\begin{align}
v(s)=\frac{d\vect{x}}{dt}(s)=v_{max}-\kappa(s)*m,s\in[0, s_{p}]
\end{align}
where $\vect{x}(s)$ is the full path as a function of $s$, $v_{max}$ is the maximum velocity for the kinematic model, $m$ is the linear kinematic mapping parameter which relates curvature to velocity, and $s_p$ is the overall path length.  We can reparametrize by time with,
\begin{align}
t=\int_{0}^{s} \frac{1}{v(s)}ds, s\in[0, s_p]
\end{align}
which yields a numerically invertible relationship between $t$ and $s$.  

To select an endpoint to feed into the trajectory optimization step, we simply evaluate the position and velocity equations at the time horizon $T_H$.

\subsubsection{Direct Trajectory Optimization}

Direct trajectory optimization formulates the nonlinear optimization problem by including both inputs and states as decision variables \cite{tedrake2009underactuated}. A collocation or transcription of the dynamics is then used to constrain the variables to ensure dynamic feasibility. For this reason, direct trajectory optimizers are well suited to accept a set of state variable initializations and are are often more robust to local minima than indirect methods \cite{tedrake2009underactuated}. Here, we use a direct transcription formulation of Simpson's integration rule as described in \cite{pardo2016evaluating} and formulate a feasibility problem--- our objective is to ensure dynamic feasibility of the trajectory and collision avoidance.  To solve our direct trajectory feasibility problem, we employ the Sparse Nonlinear Optimizer (SNOPT) \cite{Gill:2005:Snopt}. Collision avoidance is formulated as a non-penetration constraint on the occupancy grid.

Our feasibility problem can be written as
\begin{equation}
\begin{aligned}
& \underset{\vect{x}_k, \vect{u}_k, h }{\text{min}}
& & 0 \\
& \text{s.t.}
& & \forall k \in \{0,\ldots N\} \text{ and }\\
&
& & \vect{x}_{k} - \vect{x}_{k+1} + \frac{h}{6.0}(\vect{\dot{x}}_{k} + 4 \vect{\dot{x}}_{c,k} + \vect{\dot{x}}_{k+1})= 0\\
&
& & \vect{x}_f - \vectg{\delta}_f \le \vect{x}_{N} \le \vect{x}_f + \vectg{\delta}_f\\
&
& & \vect{x}_i - \vectg{\delta}_i \le \vect{x}_{N} \le \vect{x}_i + \vectg{\delta}_i\\
&
& & \vect{x}_{min}  \le \vect{x}_{k} \le \vect{x}_{max},~~ \vect{u}_{min}  \le \vect{u}_{k} \le \vect{u}_{max}\\
&
&  & d(\vect{x}) \geq r \\
&
& & h_{min}  \le h \le h_{max}
\end{aligned}
\end{equation}
where 
\begin{align}
\vect{\dot{x}}_{k} &= \vect{f}(t, \vect{x}_{k}, \vect{u}_{k}),~~
\vect{\dot{x}}_{k+1} = \vect{f}(t, \vect{x}_{k+1}, \vect{u}_{k+1})\nonumber\\
\vect{u}_{c,k} &= (\vect{u}_{k} + \vect{u}_{k+1}) / 2\nonumber\\
\vect{x}_{c,k} &= (\vect{x}_{k} + \vect{x}_{k+1}) / 2 + h (\vect{\dot{x}}_{k} - \vect{\dot{x}}_{k+1}) / 8\nonumber\\
\vect{\dot{x}}_{c,k} &= \vect{f}(t, \vect{x}_{c,k}, \vect{u}_{c,k}).
\end{align}

Here, $\vect{x}$ signifies the system state and $\vect{u}$ is the control actions. $h$ is the time step, $\vectg{\delta}_f$ and $\vectg{\delta}_i$ represent the bounds on the desired final and initial states ($\vect{x}_f$,$\vect{x}_i$), respectively. $N$ is the number of knot points. The minimum distance function $d$ is generated using the distance map from an OctoMap \cite{hornung2013octomap}.

\subsection{Local Linear Feedback Control}
To control the aircraft to a trajectory between plans, we use time-varying LQR (TVLQR).
\begin{align}
\bu(t,\bx) &= \matr{K}(\bx-\bx_0(t))+\bu_{0}(t).
\end{align}
$\matr{K}(t)$ is found by integrating
\begin{align}
-\dot{\matr{S}}(t) = \matr{A}(t)^T\matr{S}(t)+\matr{S}(t)\matr{A}(t)\ldots\nonumber\\-\matr{S}(t)\matr{B}(t)\matr{R}^{-1}\matr{B}(t)^T\matr{S}(t)+\matr{Q}
\end{align}
backwards in time from $t=T$ to $t=0$. Here $\matr{A}(t)=\frac{\partial \bff(\bx_0,\bu_0(t))}{\partial \bx}$ and $\matr{B}(t)=\frac{\partial \bff(\bx_0(t),\bu_0(t))}{\partial \bx}$ and 
\begin{align}
\matr{K}(t) = \matr{R}^{-1}\matr{B}(t)^T\matr{S}(t).
\end{align}
Here $\bx_{0}(t)$ and $\bu_{0}(t)$ signify the nominal trajectories. $\matr{Q}$ and $\matr{R}$ are the costs on state and action respectively. $\matr{S}(t)$ is defined by a final cost $\matr{Q}_f$.
\section{Real-Time Trajectory Optimization Performance}
To evaluate the performance of our NMPC approach, we explore trajectory optimization for a post-stall 90 degree turn using our fixed-wing UAV model. We consider this 90 degree turn to be a representative maneuver for aerobatic post-stall motion planning. We consider two metrics--- computation speed and trajectory following performance. Direct transcription methods, while computationally efficient (i.e. able to exploit sparsity), do not provide the same kind of accuracy as is provided by some shooting and direct pseudospectral methods. Therefore, we do not use integration error as a metric, but rather error in following a trajectory using local feedback. LQR provides this metric as a cost-to-go, and to evaluate trajectory following, we examine the final cost of a simulated vehicle following this trajectory using TVLQR. With regards to computation time, we look at the time required to find an optimal solution when a naive seed is provided (i.e. linear progression of states from initial state to the goal) and when SNOPT's warm-start feature is used with a feasible seed. To test the warm-start, we sample from a set of random-initial condition states and examine the time-required to re-optimize the trajectory. 

\begin{figure}
  \centering
  \includegraphics[width=.99\linewidth]{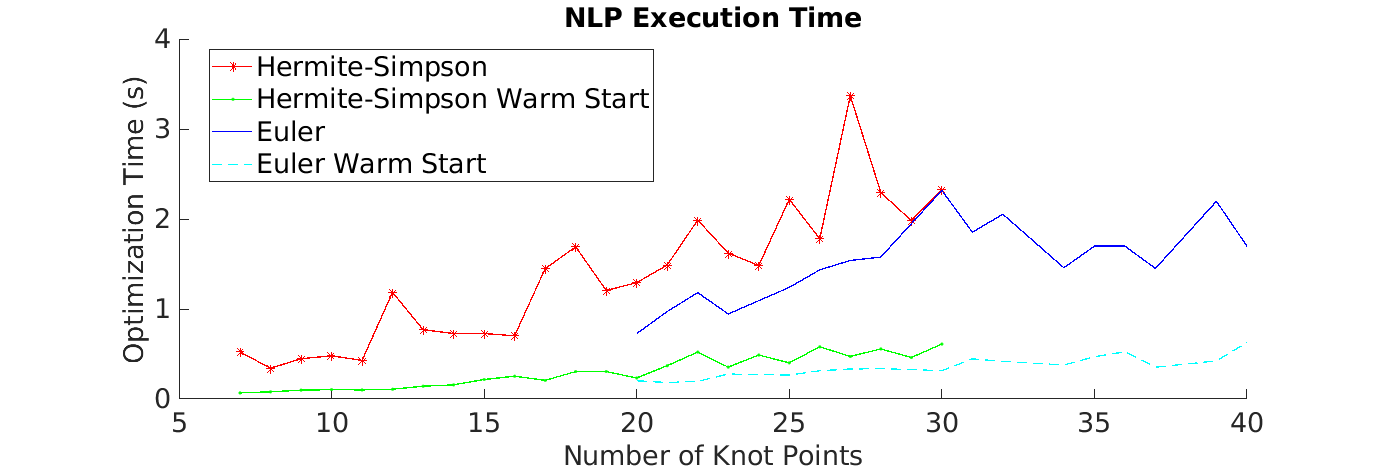}
  \vspace{-5mm}
  \caption[nlpexecution]{Plot shows time to compute an optimal trajectory using Hermite-Simpson (red star, green dot) and Euler integration methods (blue line, light blue dash). There is a significant difference between the time needed to compute a trajectory using a naive seed and using a warm-start.}
  \label{Fig:nlpexecution}
  \end{figure}

\begin{figure}
  \centering
  \includegraphics[width=.99\linewidth]{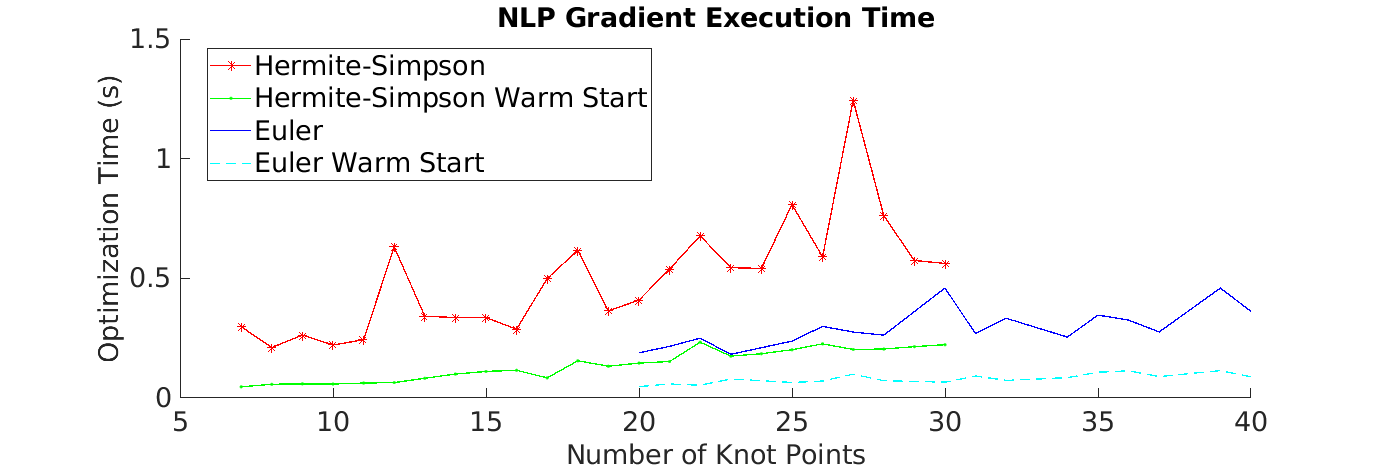}
    \vspace{-5mm}
\caption[nlpgradient]{Plot shows time spent executing the dynamics function using Hermite-Simpson (red star, green dot) and Euler integration methods (blue line, light blue dash). There is a significant difference between the time needed to compute a trajectory using a naive seed and using a warm-start.}
  \label{Fig:nlp_gradient_execution}
  \end{figure}

 
As can be observed in Figures \ref{Fig:nlpexecution} and \ref{Fig:nlp_gradient_execution}, the Euler integration scheme does not become feasible until $\approx$20 knot points are used. In general, the Hermite-Simpson integration method out-performs the Euler integration method by a factor of almost 2-1 for both the naive seed optimization and warm-start optimization, especially below 30 knot points. The Hermite-Simpson approach shows good trajectory following at 10 knot points, while the Euler integration method does not show comparable trajectory following until 24 knot points. SNOPT's warm-start also shows a 4:1 improvement over the naive seeding. 

The time spent in the gradient calls for the Euler integration warm-start is in general less than the Hermite-Simpson warm-start, which is reasonable since the Hermite-Simpson gradients are more computational intensive.

Given these results, we chose to use Hermite-Simpson integration as our integration method with 10 knot points. This provides the best computational performance ($\approx$ 0.1s for a warm-start) and reasonably good trajectory following. 

  \begin{figure}
  \centering
   \includegraphics[width=.99\linewidth]{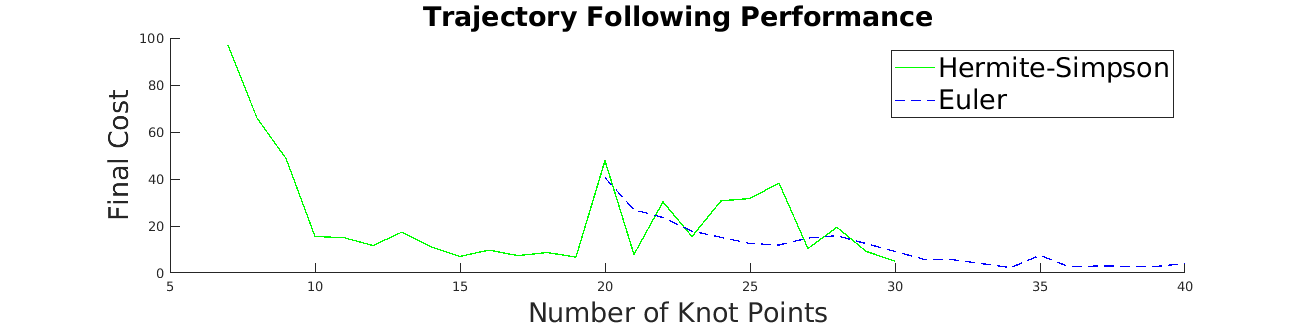}
     \vspace{-5mm}
   \caption[trajectory following]{Final cost for trajectory following of trajectories generated using Hermite-Simpson (blue) and Euler (green) integration.}
\label{Fig:trajectory_following_performance}
  \end{figure}


\section{Hardware Experiments}

\begin{figure}
  \centering
  \includegraphics[width=.5\linewidth]{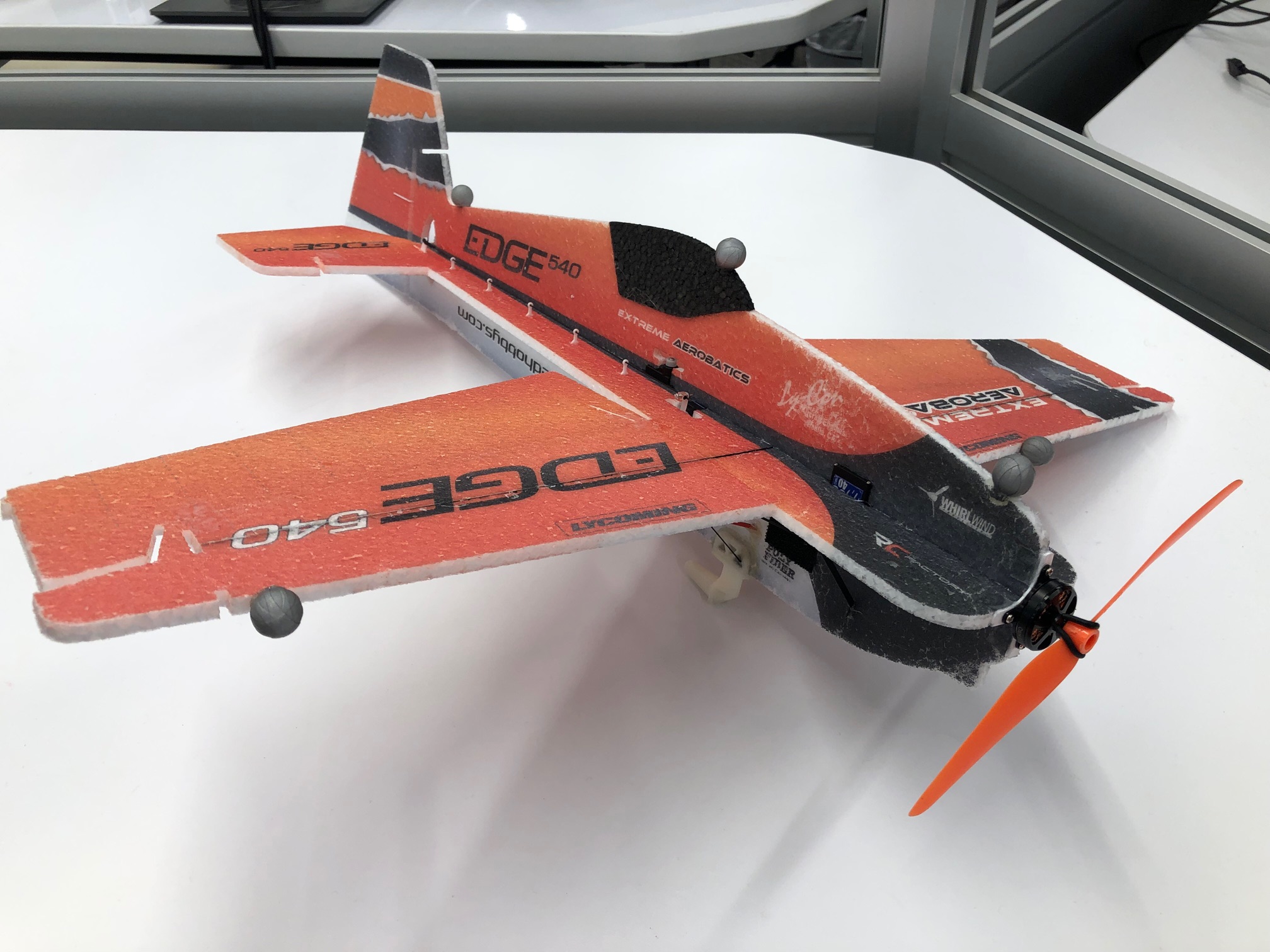}
\caption[Experimental Aircraft]{24" wingspan Edge 540 EPP model.}
  \label{Fig:timelapse}
  \end{figure}

\subsection{Experimental Set-up}
The plane used for the experiments was a 24" wingspan Edge 540 EPP model produced by Twisted Hobbys (Figure 6), with an overall weight of 120g.  It has an independently controllable rudder and elevator, as well as a pair of kinematically linked ailerons.  A 13g 2300kv Crack Series out-runner motor was paired with a 7x3.5 GWS propeller for the thruster.  In addition to the stock components for the aircraft, five Vicon markers were attached at various locations on the airframe.  A custom bungee launcher was built to produce consistent initial conditions for each trial, and an adapter designed to interface with the launcher was attached to the underside of the plane.  Maximum throws were measured for each control surface along with their corresponding PPM inputs in order to generate a linear mapping between desired control surface angle and PPM value.  

The experiments were performed in an 8m x 8m Vicon capture area, with 10 Vantage V5 cameras distributed along the periphery of the space.  The Vicon system was configured to report position and quaternion data at 200Hz to the machine running the planning and control algorithms, which was a Dell Precision 5530 laptop with an Intel i7-8850H CPU.  The planning machine used the ROS package vicon\_bridge to generate a ROS topic with the timestamped pose data, which was then then differentiated over a single timestep to obtain linear and angular velocities.  In addition, the quaternion data was transformed into yaw/pitch/roll Tait-Bryan angles.  

In order to deliver desired control inputs to the plane, a serial interface was developed to communicate with an 8MHz Arduino Pro Mini microcontroller.  The planning computer sends the desired PPM values to the microcontroller, which relays this signal to a Spectrum DX6i transmitter module through the trainer port.  The transmitter then broadcasts the signal to the RC receiver on the plane.  Since the trainer signal is only broadcast when the trainer switch is actively depressed on the transmitter, the trainer switch also acted as an emergency stop for the experiment.

\subsection{System Identification}
To account for discrepancies between the modeled dynamics and the actual plane dynamics, we compared linear and angular accelerations predicted by the dynamics model with actual measurements from flight experiments. Modification of a few parameters was found to significantly improve the accuracy of the model.  Specifically, we scaled the wing and horizontal fuselage area by a factor of 2, scaled the rudder area by a factor of 0.75, and set the backwash velocity coefficients to $\gamma_{ar}=\gamma_{al}=\gamma_{r}=0.1$, $\gamma_e=0.3$.  In addition, we estimated the linear thrust model coefficients as $a_t=-4.9167$, $b_t=9.6466$.  Results of these model adjustments can be found in Figure 7.

\subsection{Control Experiments}
The experiments were conducted with a virtual map consisting of a system of hallways roughly 1.75m wide.  A goal point was specified in the hallway directly adjacent to the initial conditions of the plane, and several variations of the experiments were conducted.  For one set of trials, the launcher was used to provide a consistent set of initial conditions, and the controller was set to execute the full receding horizon control stack.  In another variant, the stack was executed through the receding horizon trajectory optimization step, but the input trajectories were commanded open loop, without the feedback term generated from TVLQR. Additionally, several handthrown trials were performed using the full control stack.

For the constraints on the trajectory optimization problem, strong bounds on final position ($\vectg{\delta_f}(x)=\vectg{\delta_f}(y)=0.1m,\vectg{\delta_f}(z)=0.2m$) and weak bounds on final velocity ($\vectg{\delta_f}(v_x)=\vectg{\delta_f}(v_y)=3\frac{m}{s},\vectg{\delta_f}(v_z)=0.5\frac{m}{s}$) were found to produce a reasonable balance between feasibility and desired behavior.  Overall,
\[\vectg{\delta_f} = \begin{smallmatrix}[0.1,0.1,0.2,0.5,1,0.2,100,100,100,100,100,3,3,0.5,2,2,2]\end{smallmatrix}\]
The diagonals of the cost matrices $\vect{Q}$ and $\vect{R}$ for the local linear feedback controller were set to,
\begin{align}
diag^{-1}(\vect{Q}) &= [25,25,25,50,50,50,2,2,2,2,2,2,2,2,2,2,2]\nonumber\\
diag^{-1}(\vect{R}) &= [0.1,0.1,0.1,25]\nonumber
\end{align}
with final costs:
\[diag^{-1}(\vect{Q_f}) = \begin{smallmatrix}[100,100,100,100,100,100,1,1,1,1,1,1,1,1,1,1,1]\end{smallmatrix}\]
A time horizon of $T_H=1s$ was chosen as the minimum value that allows planning with sufficient forethought to execute complex maneuvers.  The replanning frequency was chosen as 5Hz based on the data from Figure \ref{Fig:nlpexecution}.  


\label{sec:experiments}

\section{Results}
The testing of our direct NMPC algorithm in hardware demonstrated very successful results. For the task where we navigated in a virtual corridor, 10 out of 10 trials navigated without colliding with our virtual walls (see green lines in Figure \ref{Fig:receding_horizon}). At the first corner, the highest AoA was 38 degrees and 66 degrees at the second corner. We also tested our algorithm without local feedback between plans, and these trials almost immediately failed (see red lines in Figure \ref{Fig:receding_horizon}). While we did occasionally violate our constraint towards the end of the trials (0.55 m radius), the actual plane never collided (see Figure \ref{Fig:bubble}). We believe this is acceptable since SNOPT is able to relax these constraints during elastic mode operation to minimize infeasibilities. To test robustness to initial conditions without a launcher, we also conducted some handthrown trials and had good success see (Figure \ref{Fig:receding_horizon}). 

  \begin{figure}
\centering
\begin{minipage}{.24\textwidth}
  \centering
  \includegraphics[width=.9\linewidth]{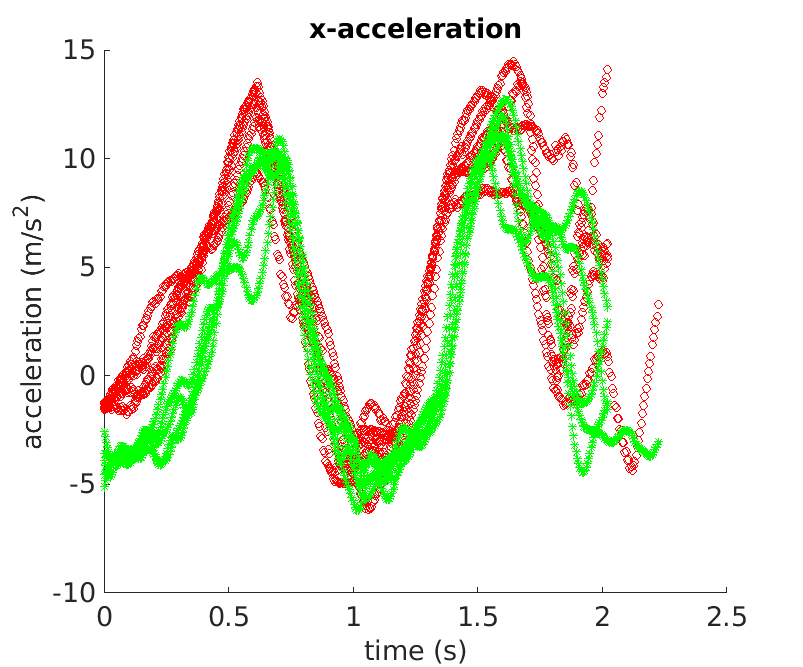}
\end{minipage}%
\begin{minipage}{.24\textwidth}
  \centering
  \includegraphics[width=.9\linewidth]{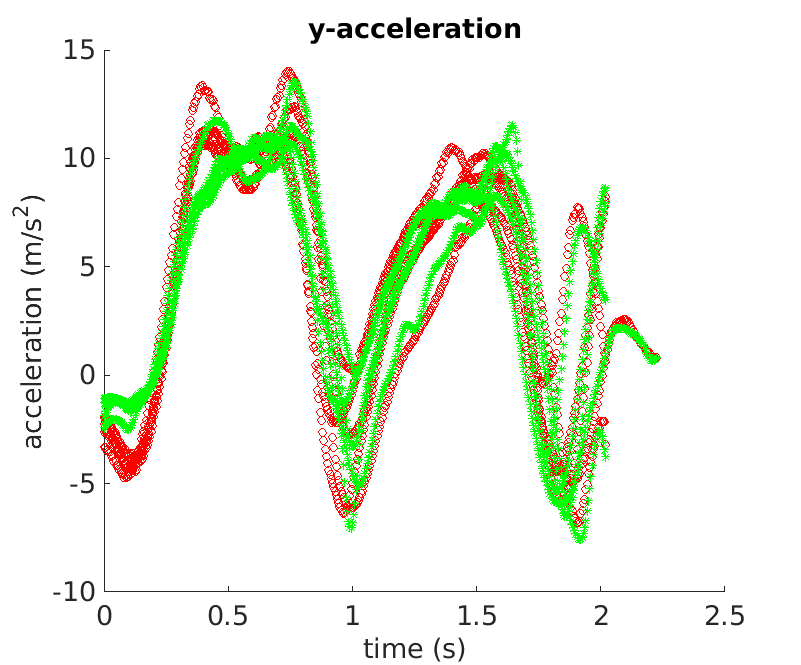}
\end{minipage}
\caption[sys id1.]{Plots show the linear x and y accelerations predicted by the model and those recorded during the control trials. Red circles are the accelerations predicted by our model. Green asterisks are the recorded data.}
  \label{Fig:sysid1}
  \end{figure}
  \begin{figure}
\begin{minipage}{.24\textwidth}
  \centering
  \includegraphics[width=.9\linewidth]{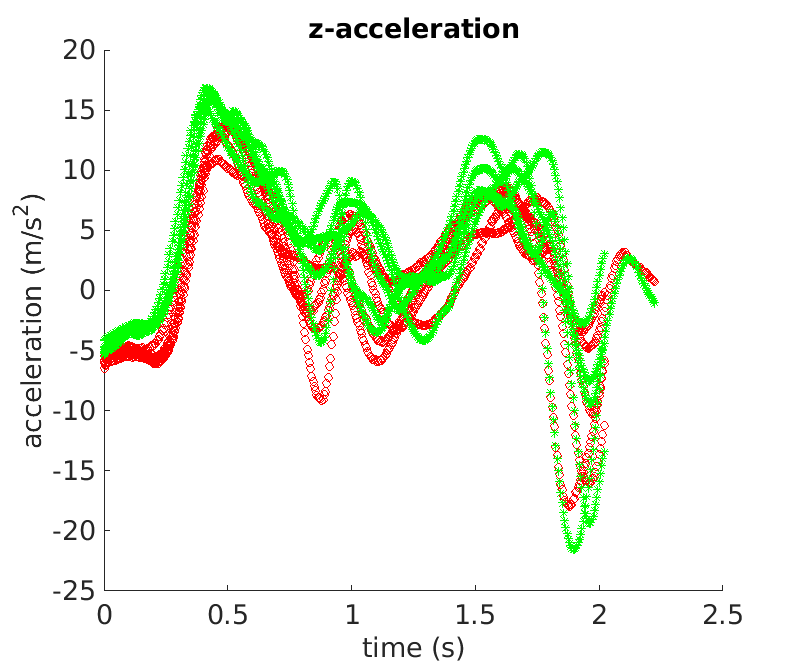}
\end{minipage}
\begin{minipage}{.24\textwidth}
  \centering
  \includegraphics[width=.9\linewidth]{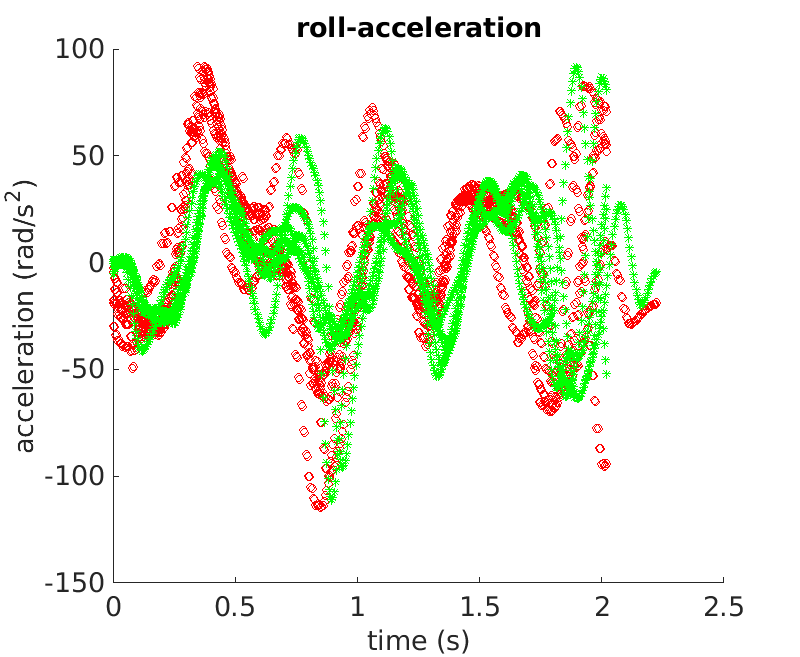}
\end{minipage}
\begin{minipage}{.24\textwidth}
  \centering
  \includegraphics[width=.9\linewidth]{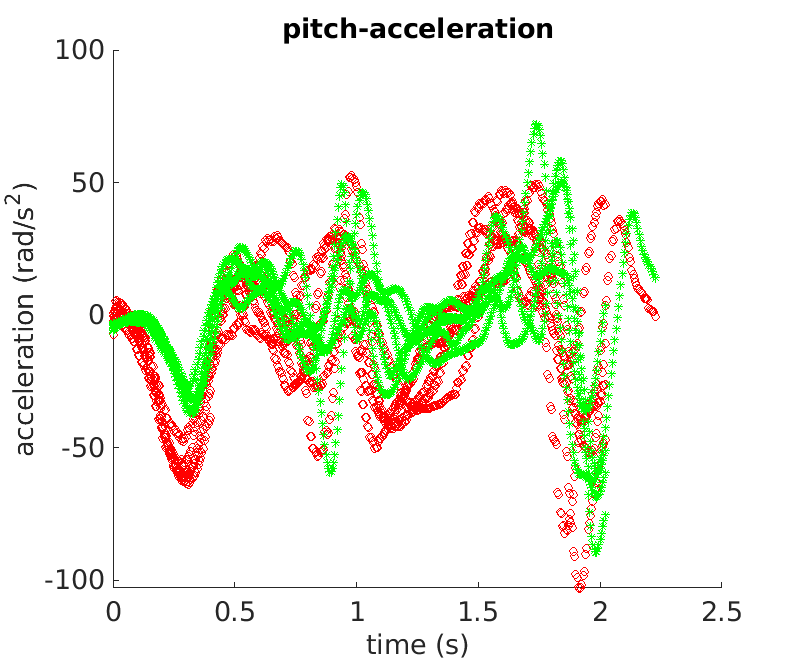}
\end{minipage}
\begin{minipage}{.24\textwidth}
  \centering
  \includegraphics[width=.9\linewidth]{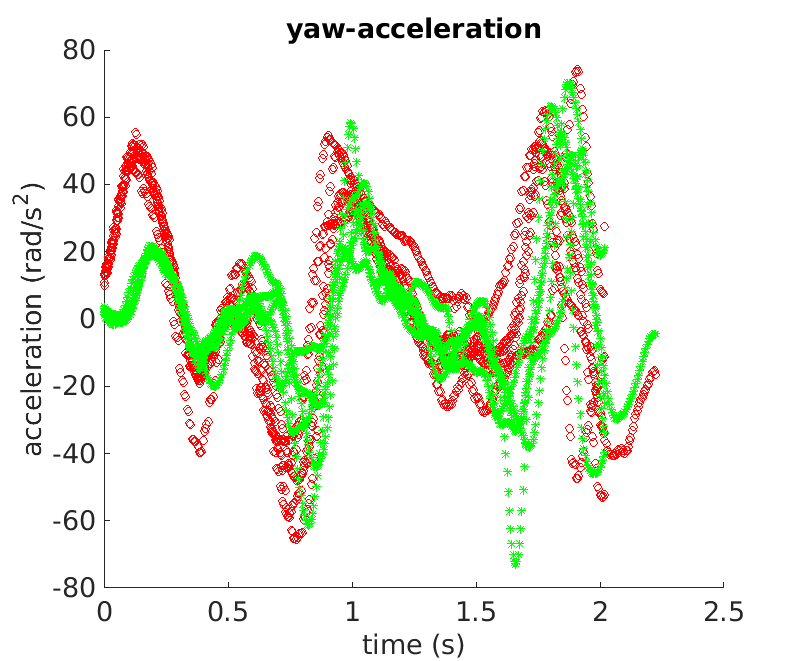}
\end{minipage}
\caption[sys id.]{Plots show the linear and angular accelerations predicted by the model and those recorded during the control trials. Red circles are the accelerations predicted by our model. Green asterisks are the recorded data.}
  \label{Fig:sysid}
\end{figure}
%
   
  \begin{figure}
  \centering
  \includegraphics[width=.8\linewidth]{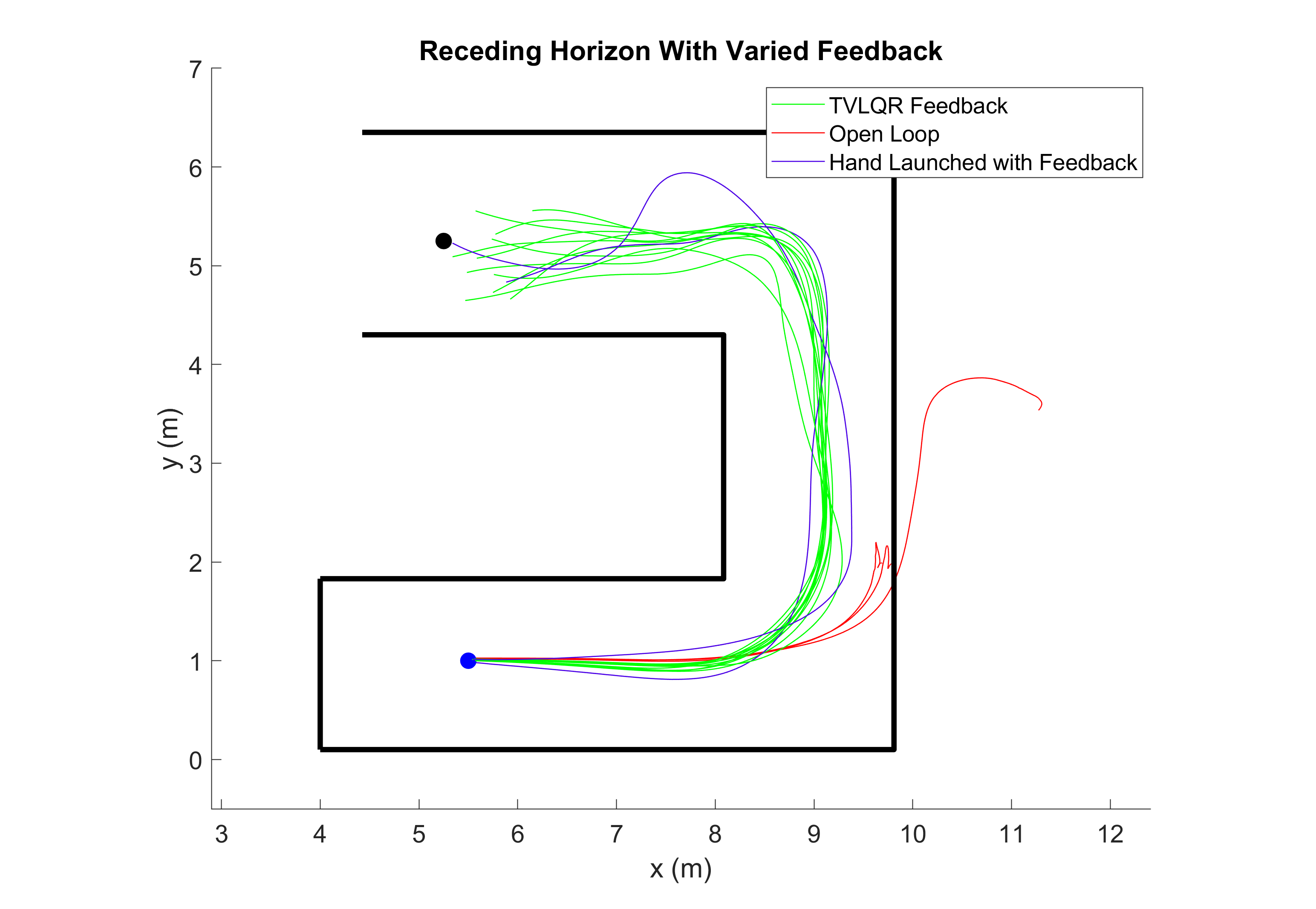}
\caption[Fixed-Goal Replanning]{Top-down plot of the receding horizon trials with (green) and without local feedback (red) and handlaunched trials (blue). All receding horizon trials with local feedback are successful while the trials without feedback fail.}
  \label{Fig:receding_horizon}
  \end{figure}
  
\begin{figure}
  \centering
  \includegraphics[width=.8\linewidth]{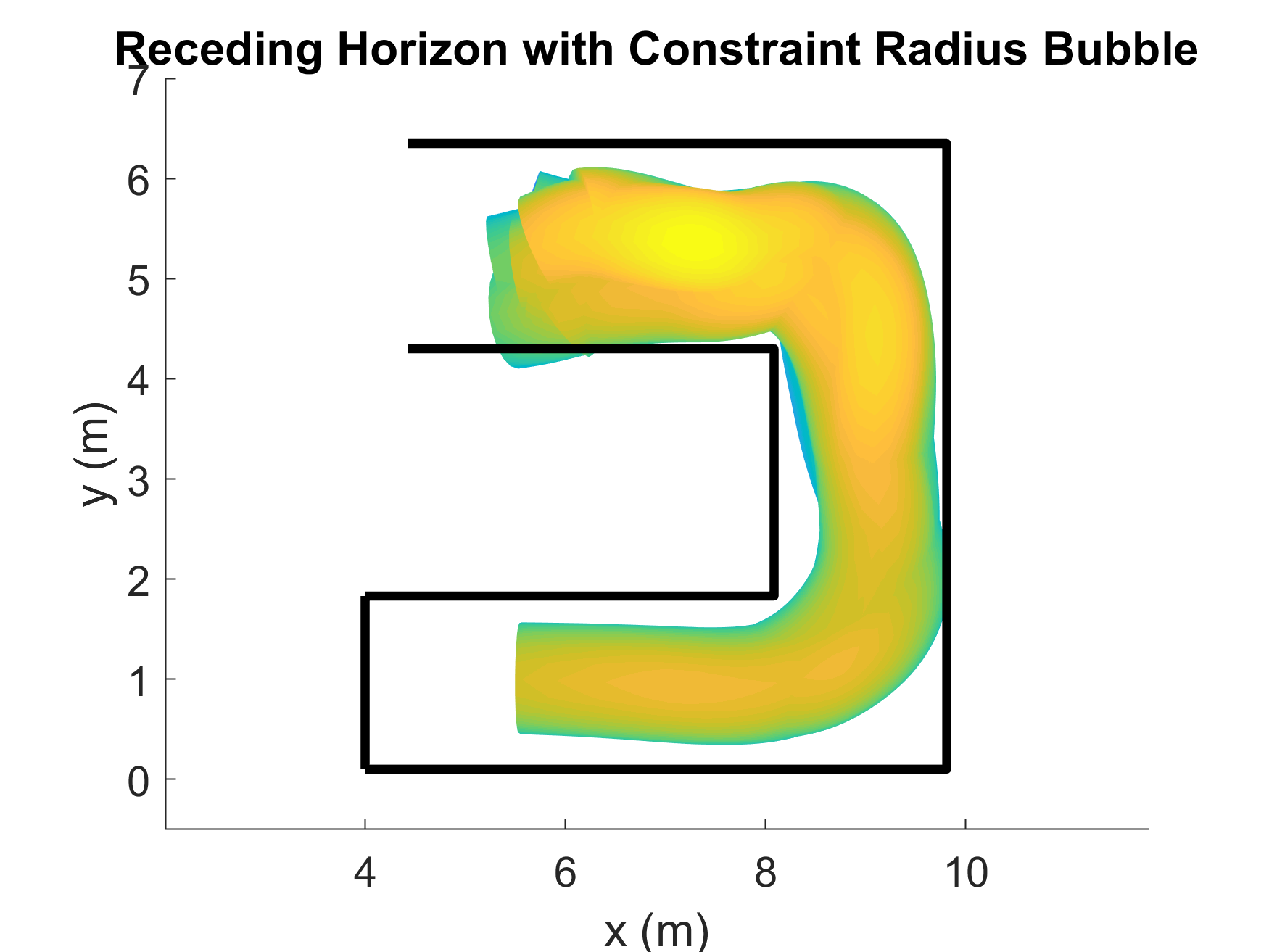}
\caption[Receding Horizon no-feedback]{Top-down plot of the spherical collision constraint from the receding horizon trials. Minor constraint violations are observed at the end of the maneuver, but the vehicle still remains collision free.}
  \label{Fig:bubble}
  \end{figure}

%
  
\label{sec:results}

\section{Discussion}
\label{sec:discussion}
In this paper, we described an approach for real-time receding horizon control with fixed-wing UAVs that can exploit post-stall aerodynamics to make tight turns in constrained environments. We demonstrated very successful experimental results with a medium-fidelity aircraft physics model. We believe that one avenue of future work is understanding how the performance of local feedback laws under uncertainty can be incorporated into the optimization routine to carry-out robust real-time feedback motion planning, such has been explored \cite{manchester2017dirtrel,desaraju2017experience}. Another avenue to explore would be the impact of sensing on these maneuvers and how real-time post-stall motion planning can coupled with perception, as has been explored in the quadcopter domain (e.g. \cite{falanga2018pampc}).


\bibliographystyle{plainnat}
\newpage
\newpage
\bibliography{references}

\end{document}